# Leading Tree in DP_CLUS and Its Impact on Building Hierarchies

Ji Xu[1,3,4], Guoyin Wang[1,2,3]✉

[1]*School of Information Science & Technology, Southwest Jiaotong University, Chengdu, China*

[2]*Chongqing Key Laboratory of Computational Intelligence, Chongqing University of Posts and Telecommunications, Chongqing, China*

[3]*Institute of Electronic Information Technology, Chongqing Institute of Green and Intelligent Technology, CAS, Chongqing, China*

[4]*Information Engineering School, Guizhou University of Engineering Science, Bijie, China*

**Abstract:** This paper reveals the tree structure as an intermediate result of "clustering by fast search and find of density peaks" (DP CLUS), and explores the power of using this tree to perform hierarchical clustering. The array used to hold the index of the nearest higher-densitied object for each object can be transformed into a Leading Tree (LT), in which each parent node P leads its child nodes to join the same cluster as P itself, and the child nodes are sorted by their g values in descendant order to accelerate the disconnecting of root in each subtree. There are two major advantages with the LT: One is dramatically reducing the running time of assigning noncenter data points to their cluster ID, because the assigning process is turned into just disconnecting the links from each center to its parent. The other is that the tree model for representing clusters is more informative. Because we can check which objects are more likely to be selected as centers in finer grained clustering, or which objects reach to its center via less jumps. Experiment results and analysis show the effectiveness and efficiency of the assigning process with an LT.

# 1. Introduction

Clustering is a general methodology and a rich conceptual and algorithmic framework for data analysis and interpretation, which gathers the objects into groups [1]. Clustering can be divided into two categories regarding the structure of the result returned. *Flat clustering* (also called *partition clustering* in some literature) returns the clusters just in one layer, which is efficient and conceptually simple but has some drawbacks. One case is that the number of clusters is just too large to make good sense in the perspective of human cognition. Because according to Miller's "seven plus or minus two" theory, there are the 7 objects in the span of attention, and the 7 digits in the span of immediate memory [2]. Too large number of clusters doesn't offer people a good understanding of the data. Another drawback is some of the real-world datasets are hierarchically structured in nature, but flat clustering is unable to reflect the truth. These limitations can be broken through by *hierarchical clustering*, which outputs a hierarchy, a structure that is more informative than flat clustering [3].

DP_CLUS is a flat clustering method able to efficiently and accurately cluster datasets of any shape with the aid of defining two simple measures: local density ρ and the distance to the nearest neighbor with higher density δ[4]. Since it was published in June, 2014, over 30 papers have been published (some are informally) to follow this research. The majority of the citing works directly adopted DP_CLUS to the problems from specific domain, such as neuroscience [5], geoscience and remote sensing [6], molecular biology [7], computational biophysics [8], age estimation in image processing [9], finding a food soulmate [10], fundamental matrix estimation in computer vision [11], analytic chemistry [12], clustering Sentences for multi-document Summarization [13], and so forth. Among the remaining part, some ensemble DP_CLUS with other methods to deal with streaming data [14] and imbalanced dataset oversampling [15], to find nonspherical clusters [16], to resolve inverse Ising problem [17], to classify scene image combined with *k* means [18], etc.; Wang discussed its parameter the cut off distance *dc* setting [19]; Zhang extended it to cluster the datasets without density peaks [20]. There are some works only used its partial idea, such as neighbor to build a *k*-nearest neighborhood to explore the clusters in high dimension and large scale datasets [21]; or just mentioned it as a clustering method [22, 23].

Among the research works mentioned above, Teng Qiu's [24] and our method appear somehow alike. But the differences are obvious and significant: **a)** Qiu constructs an In-Tree with hopping the data points to the locations of their first transfer points; but we directly transform the array ***Nn,*** which holds the index of the nearest point of higher density and are computed in DP_CLUS, into a plain tree. In this tree, each node except the root is led by its parent to join the same cluster. Thus we call this tree a *Leading Tree* (LT). **b)** Qiu evolves the In-Tree constructed into the final stars to represent the clustering result; but we remain the LT unchanged as an intermediate result permanently ready for further hierarchical clustering with arbitrary number of clusters.

In the paper, we present the algorithms of transforming the array ***Nn*** into an LT and performing hierarchical clustering based on the LT data structure. Experiment results and analysis find the boundary condition of using an LT to save time in hierarchical clustering with DP_CLUS, and show that the clustering results described with trees are more informative than the original approach. The rest of the paper is organized as follows. Section 2 reviews DP_CLUS. Section 3 presents the algorithms of transforming ***Nn*** into an LT and performing hierarchical clustering based on the LT. The experimental setting, datasets and efficiency evaluation and discussion are described in Section 4, and we draw a conclusion in Section 5.

## 2. Preliminaries

The method proposed is mainly based on [4], so we give a brief introduction to its idea and the algorithm here. The authors firstly made a sound intuitive assumption that no matter what the shape of clusters looks like, centers are always surrounded by non-center data points with lower density, and the distance between two centers are relatively long. Then two simple measures, namely local density (denoted as ρ) and minimal distance to data points with higher density

(denoted as $\delta$), are employed to accomplish the clustering job.

The notations used to describe the algorithms in DP_CLUS and our method are listed in Table 1.

TABLE 1: Notations in DP_CLUS

| Symbol | Meaning |
| --- | --- |
| $X = \{x_1, x_2, ..., x_N\}$ | The dataset with $x_i$ as its $ith$ data point |
| $D = \{d_{i,j}\}$ | The distances of the pairs of data points in $X$, where $1 \leq i < j \leq N$ |
| $I = \{1, 2, ..., N\}$ | The set of the indices of data points in the dataset |
| $\rho_i$ | The local density of $x_i$ |
| $P = (\rho_{q_1}, \rho_{q_2}, ..., \rho_{q_N})$ | The sorted local density series in descending order |
| $Q = (q_1, q_2, ..., q_N)$ | The indices of data points in $\rho$ descending order |
| $\Delta = \{\delta_1, \delta_2, ..., \delta_N\}$ | The distance to nearest points of higher local density |
| $Nn = (n_1, n_2, ..., n_N)$ | The indices of the nearest neighbor with larger $\rho$ for data points series $(x_1, x_2, ..., x_N)$ |
| $\Gamma = \{\gamma_1, \gamma_2, ..., \gamma_N\}$ | $\gamma_i = \rho_i \times \delta_i$, for $i$ in I |
| GammaSortInds | I sorted by the corresponding $\gamma$ values of data points in descendant order |

DP_CLUS takes the distance matrix of a given dataset as input, and performing the following steps:

(1) Compute $\{\rho_1, \rho_2, ..., \rho_N\}$ via cut-off kernel:

$$\rho_i = \sum_{j \in I \setminus \{i\}} \chi(d_{i,j} - d_c), \text{ where } \chi(x) = \begin{cases} 1, & x < 0; \\ 0, & x \geq 0. \end{cases} \quad (1),$$

$d_c$ is the cutoff distance; or via Gaussian kernel:

$$\rho_i = \sum_{j \in I \setminus \{i\}} e^{-(\frac{d_{i,j}}{d_c})^2} \quad (2).$$

Equation (2) is used in the authors' implementation.

(2) Sort $\{\rho_1, \rho_2, ..., \rho_N\}$ in descending order to get $P$;

(3) Compute $\Delta$ via

$$\delta_{q_i} = \begin{cases} \min_{\substack{q_j \\ j<i}} \{d_{q_i, q_j}\}, & i \geq 2; \\ \max_{j \geq 2} \{d_{q_i, q_j}\}, & i = 1. \end{cases} \quad (3),$$

and write the index of the nearest neighbor with larger $\rho$ in vector **Nn**;

(4) Interactively choose the points with "anomalously large" $\delta$ and $\rho$ as centers;

(5) Assign each data points to the same cluster as its nearest neighbor with larger $\rho$. This process is done by scanning the vector $P$ only once with referring to **Nn**.

The authors use a parameter named *bord_rho* to distinguish core and hallo data points of a cluster, but for simplicity, we omit the discussion of hallo and core in this study.

# 3. Leading Tree as an Intermediate Result of DP_CLUS

The assigning process in DP_CLUS is based on two arrays: *Nn* and *Q*. *Q* holds the indices of data points sorted by their ρ values in descending order. The whole data points are assigned to the clusters one by one in the ρ −descending order by referring to *Nn*. See Algorithm 1.

**Algorithm 1.** AssignDP_CLUS [4]
Input: *Nn* and *Q*
Output: Cluster label for every objects in the form of a array *cl*
Step 1. Initialize every elements in cl with -1;
Step 2. Label each center with a cluster id;
Step 3. For each $q_i$ in *Q* do
       if *cl*[$q_i$]==-1
          *cl*[$q_i$]:= *Nn*[$q_i$];
       End if
   End for
Step 4. return *cl*;

The computational complexity for Algorithm 1 is 2×*N*, where *N* is the number of objects.

## 3.1 Constructing Leading Tree

The LT of a give distance matrix can be constructed by directly transforming the array *Nn* and another array named *GammaSortInds*. Since Nn indicates its direct leading parent *P* for each object, then it is easy to find out the child nodes led by *P*. By adding the corresponding child nodes to all parent nodes, an LT is constructed. See Algorithm 2. Note that the child nodes of each parent are added in the γ- descending order, so that when a newly upgrade center is detached from its parent, we simply remove the links from the parent to its first child. Thus the construction of the LT is accelerated. See Algorithm 2.

**Algorithm 2.** Transforming the *Nn* into an LT
Input: *Nn* and *SortedGammaInds*
Output: An LT in the form of adjacent list *AL*
Step 1. Initialize the adjacent List for each object;
Step 2. For *i* from 2 to N
        ChildID:= *SortedGammaInds*[*i*];
        ParentID:= *Nneigh*[ChildID];
        *AL*[ParentID].add(ChildID);
    End For
Step 3. return *AL*;

The computational complexity for Algorithm 2 is 3×N, where *N* is the number of objects, and the data accessing on a linked list is less efficient than on an array.

## 3.2 Using Leading Tree to build clustering Hierarchy

From the angle of center sets, hierarchical clustering can be regarded as a series of flat clustering taking different sets of objects as its centers. So, in this section, we only discuss the method of building hierarchy of clusters with an LT and each given set of centers on a corresponding layer. How to choose the centers to form a layer in the hierarchy is beyond the scope of this paper.

With an LT constructed, the assigning process is turned into just disconnecting the *m*-1 links from each center to its parent. See Algorithm 3.

**Algorithm 3.** Split the LT
Input: An LT in the form of an adjacency list *AL*, *Nn*, and array of *m* centers *C* sorted by gamma
value in descending order
Output: A forest to represent the clustering result
Step 1. For *i*=2 to *m*
    root= *C* [*i*];
    parentID= *Nneigh*[root];
    *AL* [parentID].RemoveFirst();
End For
Step 2. return *AL*;

The computational complexity of Algorithm 3 is 3×(m-1), where m is the number of centers.

## 3.3 Example

We sample the longitude and latitude of 13 cities in north China to form an illustrative dataset (see Fig. 1), to illustrate the process and result of using LT to represent the CP_CLUS intermediate result and to cluster data points based on the constructed LT.

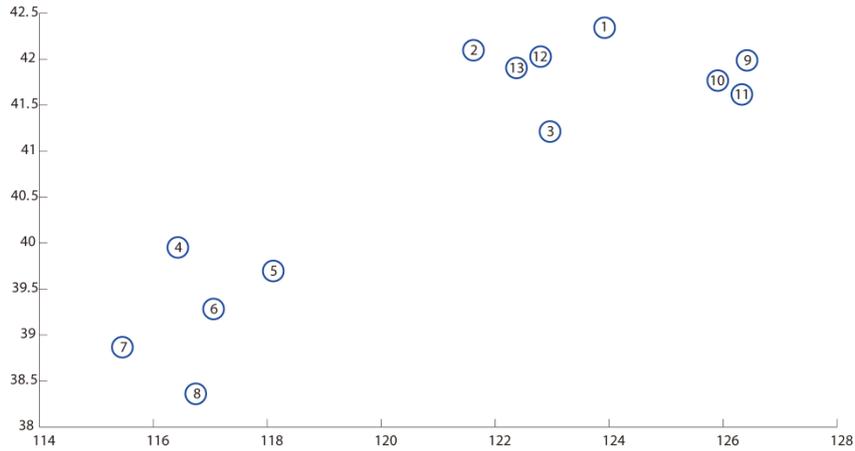
Fig. 1 Illustrative dataset of 13objects

Using the algorithm of DB_CLUS, the intermediate results *Nn*, *Q*, and *SortGammaInd* for DS1 as well as the final result *cl* for DS1 are computed, as shown in Table 2.

TABLE 2: The intermediate results and final result of DB_CLUS for DS1

| *Nneigh* | 12 | 13 | 12 | 6 | 6 | 13 | 8 | 6 | 11 | 11 | 12 | 13 | 0 |
|---|---|---|---|---|---|---|---|---|---|---|---|---|---|
| *ordRho* | 13 | 12 | 11 | 10 | 9 | 6 | 3 | 2 | 4 | 8 | 1 | 7 | 5 |
| *SortGammaInd* | 13 | 6 | 11 | 3 | 12 | 1 | 8 | 4 | 2 | 7 | 10 | 5 | 9 |
| *cl* | 1 | 1 | 1 | 2 | 2 | 2 | 2 | 2 | 3 | 3 | 3 | 1 | 1 |

In the LT approach, Nn of DS1 is transformed into to a tree with referring to SortGammaInd (Fig. 2a), and the result of clustering taking {13, 6, 11} as centers is presented in the form of a forest (Fig. 2b).

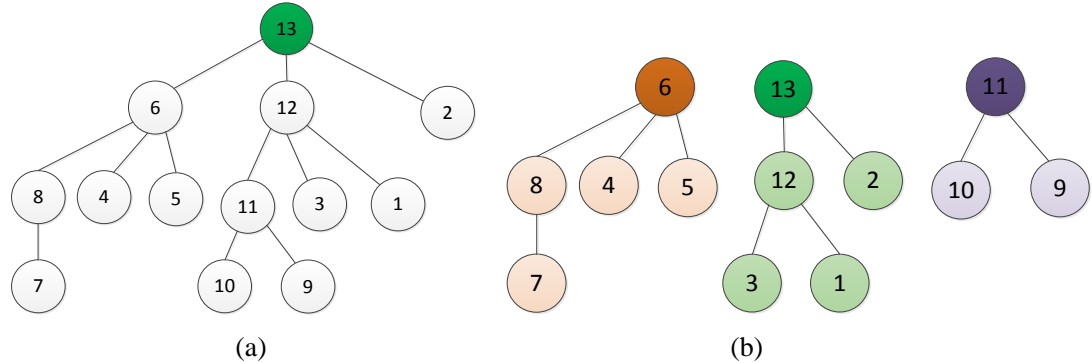

Fig.2.(a) The intermediate result of DS1 in the form of an LT, (b) Taking {13, 6, 11} as centers, the LT for DS1 is split into a forest. Each new tree represents a cluster taking the root as its center.

The physical structures to implement the logical structure of the LT and the split forest for DS1 are shown in Fig 3a and Fig. 3b respectively.

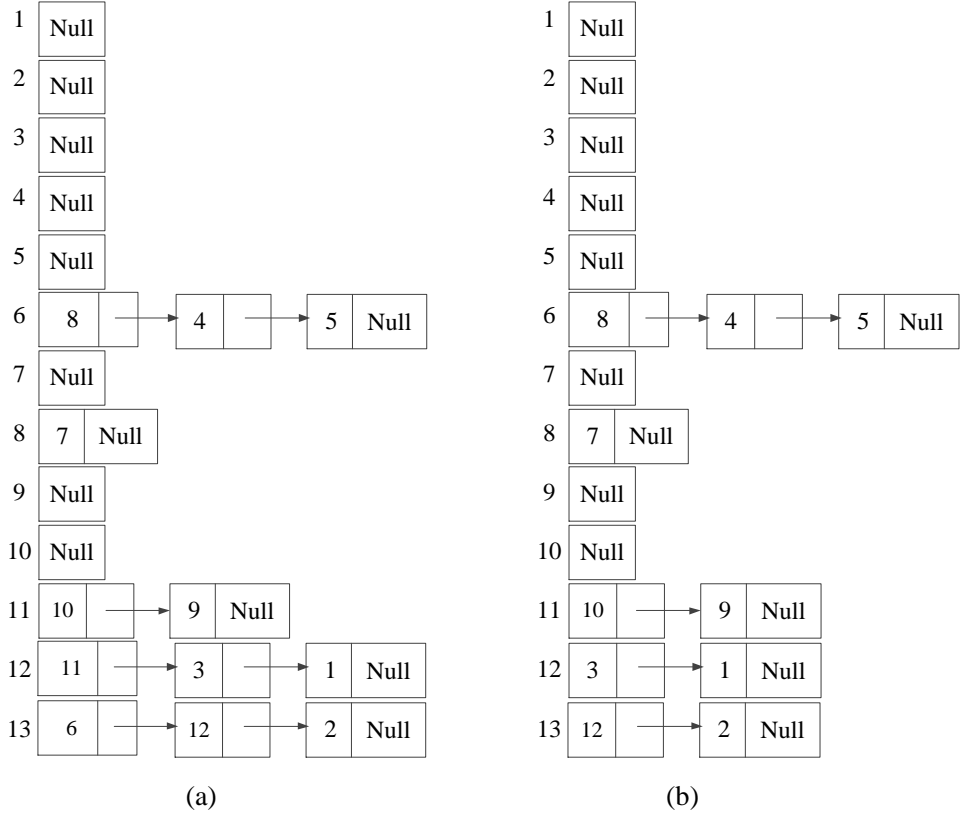

Fig. 3. physical structures to implement the logical structure of the LT (a) and the split forest (b) for DS1

The process of retrieve the cluster objects may be a litter slow than sequential table, but it is more informative. Unlike with the array cl, we can check which objects are more likely to be selected as centers in finer grained clustering, or which objects reach to its center via less jumps with the trees representing the clusters.

## 4. Experiments

### 4.1 Datasets and running settings

The experiments are conducted on a personal computer with Intel i5-2430M CPU,8G RAM, Windows 7 64bit OS, and Eclipse programming environment with JDK 1.7.

We test our method on 3 datasets: two of them are synthetic and one real world dataset from UCI Machine Learning Repository. The first dataset (5Spherical) is generated by setting centers of five spheres and randomly sampling dots on the surface of the spheres, and then projecting the dots to the plane (see Fig. 4a). 5Spherical is designed to show that the whole dataset maybe clustered as five, four or two groups. The second dataset (5Spiral) are 5 spiral curves using Function (4):

$$\begin{cases} x = -t/8 \times \cos(t+\theta) \\ y = -t/8 \times \sin(t+\theta) \end{cases} \quad \dots\dots\dots\dots\dots\dots\dots\dots\dots\dots\dots\dots\dots\dots\dots\dots\dots\dots\dots(4)$$

where $t \in (2, 4\pi)$, and $\theta$ is the parameter controls the start point of the spiral curves. Among the five spirals, two spirals in two pairs are arranged relatively close to each other and one spiral is separated, thus they can be clustered into five, three or two groups (see Fig. 4b).The two artificial datasets are both of hierarchical structure, and of spherical and nonspherical shape respectively, with which the effectiveness and robustness of a hierarchical clustering method can be tested.

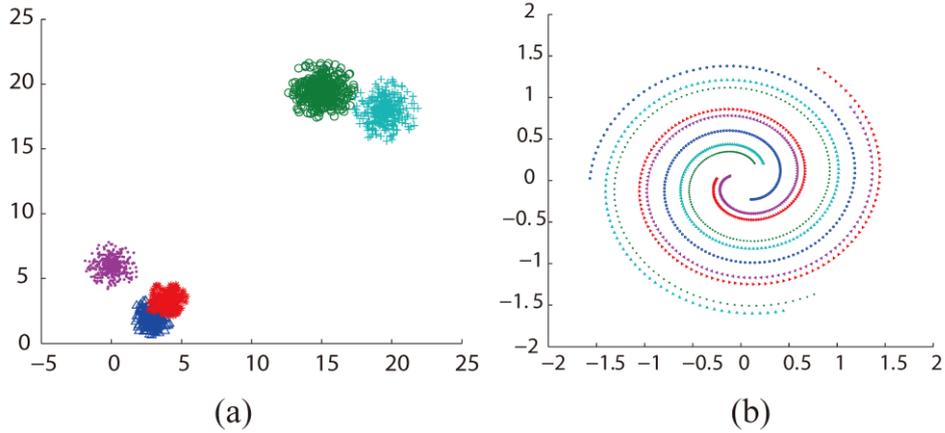

(a)　　　　　　　　　　(b)

Fig. 4: Two artificial datasets of hierarchy nature

The brief information of the 6 datasets is tabulated in Table 3.

TABLE 3: Datasets used to test LT

| Dataset | Object Number | Attribute Number | Origin |
| --- | --- | --- | --- |
| 5Spherical | 2200 | 5, 4, 2 | Artificial |
| 5Spiral | 1060 | 5, 3, 2 | Artificial |
| Ecoli | 336 | 8, 4, 2 | Real-world |

## 4.2 Results and discussion

The running time of assigning objects to cluster with Nn, splitting the LTs to obtain the clusters and constructing the LTs is depicted in Fig. 5.

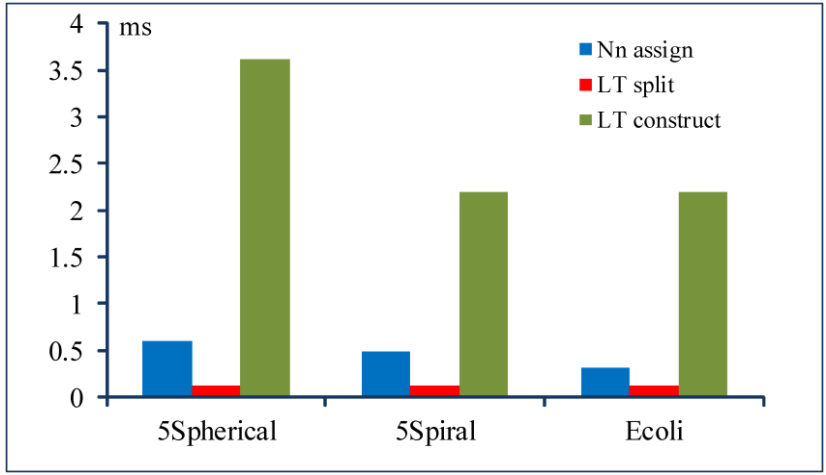

Fig. 5: Running time of the 3 algorithms on the 3 datasets

From the figure, we can see that the running time of split an LT into a forest $T_{SplitLT}$ is always much shorter than that of assigning with **Nn** (denoted as $T_{Assign}$). But the cost is the time for constructing the LT ($T_{ConstrLT}$). If the number of the potential layers in the hierarchy $Nl$ is small like the datasets we tested, then $T_{ConstrLT} + Nl \times T_{SplitLT} > Nl \times T_{Assign}$. However, if $Nl$ is so large that Equation 5 is satisfied, then using an LT to perform hierarchical clustering will save computing time.

$$Nl > \frac{T_{ConsLT}}{T_{Assign} - T_{SplitLT}} \qquad (5)$$

Besides, with the clusters being described by trees, users can tell the differences among the objects within the same cluster. For example, as shown in Fig 2.b we can say that object No.2 is closer to the center than No.1 and No.3, because there is an object (No.12) separating them from the center. This is more informative than the original approach of DP CLUS, which treat each objects in a cluster equally, just as most of the clustering methods.

## 5. Conclusion

In this paper, we reveal the hidden tree structure of the intermediate result in DP_CLUS, and develop the algorithms to construct the leading tree (LT) and to perform hierarchical clustering with a series of center sets with the LT. Both theoretical analysis and experimental results show that the LT approach is much more efficient to assign the noncenter objects in DP_CLUS than the original **Nn** approach, and the clusters presented with trees can provide more information on the noncenters' potential to be selected as centers in future and on how many jumps a given object needs to reach its center. This discovery not only holds the potential to accelerate the assigning processing during the hierarchical clustering with DP_CLUS, but also provides us with a more profound understanding on the result structure of DP_CLUS.

# Acknowledgement

This work is partly supported by National Science and Technology Major Project (NO.2014ZX07104-006), the National Natural Science Foundation of China under Grant numbers of 61272060 and 61472056, and Natural Science Foundation Key Project of Chongqing of P. R. China under Grant No. CSTC2013jjB40003.

**Ji Xu** received the B.S. from Beijing Jiaotong University, Beijing, in 2004, and the M.S. from Tianjin Normal University, Tianjin, China, in 2008. Now he is a Ph.D. candidate with Southwest Jiaotong University, Chengdu, China. His research interests include data mining, granular computing and software engineering.

**Guoyin Wang** received the B.S., M.S., and Ph.D. degrees from Xi'an Jiaotong University, Xi'an, China, in 1992, 1994, and 1996, respectively. He worked at the University of North Texas, USA, and the University of Regina, Canada, as a visiting scholar during 1998-1999. Since 1996, he has been working at the Chongqing University of Posts and Telecommunications, where he is currently a professor, the Director of the Chongqing Key Laboratory of Computational Intelligence, and the Dean of the College of Computer Science and Technology. He was appointed as the Director of the Institute of Electronic Information Technology, Chongqing Institute of Green and Intelligent Technology, CAS, China, in 2011. He is the author of 10 books, the editor of dozens of proceedings of international and national conferences, and has over 200 reviewed research publications. His research interests include rough set, granular computing, knowledge technology, data mining, neural network, cognitive computing.